\RequirePackage{amsmath}
\documentclass[runningheads]{llncs}
\usepackage{amsmath}
\usepackage{amsfonts}
\usepackage[T1]{fontenc}
\usepackage{array}
\usepackage[figuresright]{rotating}
\usepackage{graphicx}
\usepackage{booktabs}
\usepackage{longtable}
\usepackage{caption}
\usepackage{subcaption}
\usepackage{csvsimple}
\usepackage{colortbl}%
\usepackage{hyperref}
\newcommand{\myrowcolor}{\rowcolor[gray]{0.95}}

\makeatletter
\def\FPS@rot{\rotatebox{90}}
\makeatother

\usepackage{tikz}
\usetikzlibrary{trees,positioning}
\usetikzlibrary{arrows.meta}
\usetikzlibrary{decorations.pathmorphing}

\usepackage{xspace}
\usepackage{cite}

\usepackage[most]{tcolorbox}
\usepackage{listings}
\usepackage{xcolor}
\definecolor{lightgray}{gray}{0.9}
\definecolor{codeblue}{rgb}{0.2,0.2,0.6}
\definecolor{codegreen}{rgb}{0,0.6,0}
\definecolor{codered}{rgb}{0.6,0,0}
\tcbset{colback=lightgray, colframe=lightgray, sharp corners, rounded corners, boxrule=0pt, left=5pt, right=5pt, top=-5pt, bottom=-5pt}
\newtcblisting{pythoncode}{
    listing only,
    listing options={
        language=Python,
        basicstyle=\ttfamily\small,
        keywordstyle=\color{codeblue},
        commentstyle=\color{codegreen},
        stringstyle=\color{codered},
        showspaces=false,
        showstringspaces=false,
    }
}

\usepackage[colorinlistoftodos,prependcaption]{todonotes}

\newcommand{\eos}{\ensuremath{\mathsf{stop}}\xspace}

\newcommand{\Pred}{\ensuremath{p}\xspace}

\newcommand{\Act}{\ensuremath{\Sigma}\xspace}
\newcommand{\Acteos}{\ensuremath{{\Sigma_\eos}}\xspace}
\renewcommand{\epsilon}{\ensuremath{\varepsilon}\xspace}

\newcommand{\CaseIDs}{\ensuremath{\mathit{CaseID}}\xspace}
\newcommand{\caseid}{\ensuremath{\mathit{id}}\xspace}

\newcommand{\PDFA}{\mathcal{A}}

\newcommand{\act}{a}

\newcommand{\figref}[1]{Figure \ref{#1}}

\begin{document}
\title{A Framework for Streaming Event-Log Prediction in Business Processes%
\thanks{The work was supported by the French National Research Agency (ANR) projects DREAMY (ANR-21-CE48-0003) and COSTXPRESS (ANR-23-CE45-0013), as well as the SAIF project, funded by the ``France 2030'' government investment plan managed by ANR, under the reference ANR-23-PEIA-0006.}}
%
%
\author{Benedikt Bollig\inst{1}\orcidID{0000-0003-0985-6115} \and
Matthias F{\"u}gger\inst{1}\orcidID{0000-0001-5765-0301} \and
Thomas Nowak\inst{1,2}\orcidID{0000-0003-1690-9342}}
\authorrunning{B.~Bollig, M.~F{\"u}gger, and T.~Nowak}
%
\institute{Université Paris-Saclay,  CNRS, ENS Paris-Saclay, LMF, Gif-sur-Yvette, France
\and
Institut Universitaire de France, Paris, France}
\maketitle              
\begin{abstract}
We present a Python-based framework for event-log prediction in streaming mode, enabling predictions while data is being generated by a business process. The framework allows for easy integration of streaming algorithms, including language models like $n$-grams and LSTMs, and for combining these predictors using ensemble methods.

Using our framework, we conducted experiments on various well-known process-mining data sets and compared classical batch with streaming mode. Though, in batch mode, LSTMs generally achieve the best performance, there is often an $n$-gram whose accuracy comes very close. Combining basic models in ensemble methods can even outperform LSTMs. The value of basic models with respect to LSTMs becomes even more apparent in streaming mode, where LSTMs generally lack accuracy in the early stages of a prediction run, while basic methods make sensible predictions immediately.

\end{abstract}

\section{Introduction}

Digital data is omnipresent and continuously produced in today's world.
A particular type of data that is available to many organizations and institutions, either through explicit mining or implicitly within business processes, are event logs, i.e., timed traces of events recorded along the execution of a process.
Examples are examinations of patients entering a hospital, server log files, and clicks of a user browsing a website.

While such event logs are often readily available at organizations, process models are typically absent.
More so, quantitative models that allow one to describe, analyze, optimize, and predict possible future events are challenging to obtain.

The discipline of process mining~\cite{Aalst16,PMH2022} aims at automatically discovering a model of the underlying process, reflecting its
causalities and concurrency, with applications in streamlining it and finding potential bottlenecks.
Classically, process mining is applied as a batch process to post-mortem data, with the discovery process being re-executed as more data is accumulated, and properties of the model being analyzed for potential optimization~\cite{reinkemeyer2020process}.

More recently, process mining has focused on prediction tasks~\cite{van2010beyond}. In this context, a (language) model predicts the next activity of a given case based on its historical event data.
Within the realm of event-log prediction, one distinguishes between two paradigms: batch learning and streaming learning, also referred to as offline and online learning.
The choice of these approaches depends on the application as well as the kind of available data:
In batch mode, a model is trained on post-mortem data; potentially including data that is different from what is seen in production: data may be annotated post-mortem with additional information.
On the other side, streaming mode differs from batch mode in that:
(i) Events are given one by one, interleaved across cases.
(ii) Training is conducted online, i.e., in production.
(iii) Predictions are made on pre-mortem data and can be exploited for decision-making in the running process, impacting and optimizing future trajectories.
A major difference is that, in streaming mode, meaningful predictions are required early on when data is still sparse.

\paragraph{Activity prediction in discrete, case-based event logs.}
While process event logs can be diverse and include attributes of various types, in this work we consider them to consist of finitely many discrete activities, each belonging to a so called case. That is, an event consists of a case ID and an activity.
An example from healthcare would be patients as cases and organizational, diagnostic, and treatment measures as activities.
In the benchmarks used in this work, the number of activities ranges from 13 to 42.
In the streaming case, however, the set of activities does not need to be known in advance.

Given a prefix of such an event log, we study the problem of predicting the most likely next activity for a case, including a distinguished \texttt{stop} symbol to signal the end of the case.
A solution to this problem is a prediction function, also referred to as language model, which, given the prefix and the case, returns the predicted next activity.
For simplicity, we only consider prediction functions whose output for a case depends solely on the historical event data for that specific case.
For example, such functions do not capture global decisions within a hospital of adapting a diagnostic measure for a patient due to resource-limitations induced by other patients.

While various synthetic datasets have been used as benchmarks for learning algorithms for inferring probabilistic automata \cite{VerwerEH14}, it is difficult to draw conclusions from these performance metrics for real-life scenarios.
Generating meaningful patterns randomly is already challenging when considering only single-case behaviors and it is not clear how the traces of several cases should be interleaved.
By contrast, in real-life datasets, interleavings are often imposed by resource constraints or external factors, which may follow non-trivial patterns.
We thus perform experiments based on 7 well known real-world process-mining datasets \cite{bpi_challenges2020} from different domains, including healthcare, finance, and IT service management.
The number of events ranges from 15,214 to 2,514,266.
These datasets have been previously used in \cite{HinkkaLH19} to establish experimental performance metrics for recurrent neural networks.
Our results align with theirs in the context of neural networks, although the exact accuracies differ slightly.

\paragraph{Contributions.}
Our contributions are twofold.
(i) We provide a Python framework that facilitates experiments for predictions in batch and streaming process mining.
The framework allows base models to be easily assembled into composed models, which can themselves serve as components of other models, in both streaming and batch mode.
(ii) Experiments conducted in our framework demonstrate the potential of simple base language models, especially when combined in ensemble methods.
For all tested datasets, long short-term memory (LSTM) networks \cite{hochreiter1997long} outperform base models such as prefix trees and $n$-grams. However, $n$-grams with a relatively small $n$ often comes close to the LSTM's performance.
The performance gap is further reduced (or even inverted) by combining base models using ensemble methods.

\paragraph{Related Work.}
Our work touches upon several fields including automata learning and grammatical inference \cite{delaHiguera2010,Vaandrager17}, process mining \cite{Aalst16}, and various contexts within machine learning.

Compared to classical automata learning in batch mode, grammatical inference for streaming data has received considerably less attention. Notable exceptions include \cite{BalleCG14,SchmidtK14,BaumgartnerV23}, which propose efficient algorithms for probabilistic automata to address computational complexities in streaming settings. Non-quantitative incremental automata learning was explored in \cite{Dupont96}.

Event-log predictions in process mining is an active area of research. Recent works include \cite{PolatoSBL18,PegoraroUGA21,CeciLFCM14,HinkkaLH19,BreukerMDB16,AalstSS11}.
Process mining in a streaming setting has been studied, e.g., in \cite{BurattinSA14,Burattin22,KrawczykC18,ZelstDA18}. Note that
\cite{KrawczykC18} suggests using ensemble methods in the presence of noise,
though not resorting to $n$-grams.

The use of $n$-grams dates back to \cite{shannon1948mathematical} and has been widely applied to a multitude of problems including business processes \cite{BreukerMDB16}.

Apart from prediction, another interesting application domain of automata learning is verification \cite{MaoCJNLN16,MayrYCPV23,Vaandrager17,Leucker06}.
For example, \cite{MaoCJNLN16} focuses on learning deterministic probabilistic automata from batches, building on variants of the Alergia algorithm~\cite{carrasco1994learning}.
However, targeting reactive systems, the focus of this work is on infinite words rather than making predictions based on finite historical data.

Finally, several general-purpose automata learning libraries have been developed, including \cite{VerwerH17,MuskardinAPPT22,BolligKKLNP10,IsbernerHS15}.

\paragraph{Outline.}
We provide the technical background of the framework and the language models used in this paper in Section~\ref{sec:preliminaries}: prefix trees, $n$-grams, probabilistic deterministic finite automata \cite{VidalTHCC05}, and LSTMs \cite{hochreiter1997long}.
The computational Python framework is introduced in Section~\ref{sec:framework}.
The experimental setup and results are presented in Section~\ref{sec:experiments}.
We conclude in Section~\ref{sec:conclusion}, discussing potential future directions.

\section{Language Models and Ensemble Learning}
\label{sec:preliminaries}

We start with a unified, automata-based view of various base language models, including frequency prefix trees \cite{delaHiguera2010}, $n$-grams \cite{shannon1948mathematical}, and probabilistic deterministic finite automata \cite{VidalTHCC05}.
This not only allows one to compare the above approaches within a single setting but has also been adapted within our Python framework, where concrete language models are obtained by refining basic automata provided by the framework.

\subsection{Base Language Models}

We fix a nonempty finite set $\Act$ of activities, also referred to as activities.
The set is not necessarily known in advance to a language model, though an algorithm (e.g., a neural network) may require specifying an upper bound on its size (as part of the embedding dimension in case of the neural network).
We write $\Act^\ast$ for the set of finite sequences of activities.
In particular, the set includes the empty sequence $\epsilon$.
Activities are referred to as $a, b$, etc.
We denote by \eos a symbol that is not contained in $\Act$ and that is used by a language model to signal the end of a sequence.
The set $\Acteos = \Act \cup \{\eos\}$ contains all activities and the stop symbol.
In the following, let $\sigma, \tau$ be from $\Acteos$.
Activities are associated to cases.
We assume that case IDs are taken from a countably infinite set $\CaseIDs$.
Specifically, in our benchmarks, case IDs will be from the set of strings or the set of natural numbers.

A language model can take several forms.
Examples are models based on $n$-grams, bags, probabilistic automata, and recurrent neural networks.
In essence, every such language model defines a \emph{probabilistic prediction function} or, simply, prediction function $\Pred \colon \Sigma^\ast \to \Delta(\Acteos)$, where
$\Delta(\Acteos)$ is the set of probability distributions over $\Acteos$.
Given a sequence $w \in \Act^\ast$, applying~$\Pred$ yields a probability distribution over the set of possible next activities, including \eos.
In analogy to conditional probabilities, one usually writes
$\Pred(\sigma \mid w)$ for the probability $\Pred(w)(\sigma)$.
The focus in this work is on predicting the most likely next activity, which is readily obtained from a probabilistic prediction function. Note that, under certain conditions, the prediction function also allows one to generate sequences according to a probability distribution over $\Act^\ast$.

As previously stated, we follow an automata-based approach to obtain prediction functions.
This is motivated by two reasons: (i) It unifies widely-used approaches, and (ii) its formalism is close to streaming learning with incremental state updates, allowing us to compare batch and streaming mode within a single framework.
Given that state updates due to learning will be deterministic, we will use a deterministic automaton to describe these.
Following \cite{MayrYCPV23}, we define:

\begin{definition}
A \emph{probabilistic deterministic finite automaton} (PDFA) is composed of
  a finite set~$S$ of states with a dedicated initial state $s_0\in S$,
  a partial deterministic transition function $\delta\colon S \times \Act \to S$ (also called update function), and
  a mapping $\pi\colon S \to \Delta(\Acteos)$ assigning a probability distribution to every state.
\end{definition}

With a PDFA $\PDFA$, we associate a prediction function $\Pred$ inductively as follows.
We extend the automaton's transition function $\delta$ to
$\hat\delta\colon S \times \Act^\ast \to S$
over sequences in the usual way by letting $\hat\delta(s, \varepsilon) = s$
and $\hat\delta(s, aw) = \hat\delta(\delta(s, a), w)$
for all $s \in S$, $a \in \Act$, and $w \in \Act^\ast$.
The \emph{prediction for $w \in \Act^\ast$} is then given by $\Pred(w) = \pi(\hat\delta(s_0,w))$.
\figref{fig:3gram-pdfa} depicts a PDFA. Its initial state is the topmost state. We have, e.g.,
$p(\eos \mid \varepsilon) = 0$,
$p(\eos \mid aa) = \tfrac{7}{13}$,
$p(a \mid aa) = \tfrac{5}{13}$, and
$p(b \mid aa) = \tfrac{1}{13}$.

Towards a data structure that provides natural ways to update transition functions, we
  use a related concept of frequencies rather than probabilities on activities \cite{delaHiguera2010}:
A \emph{frequency deterministic finite automaton} (FDFA) is defined analogously
  to a probabilistic deterministic automaton, the only difference being that, instead
  of a mapping $\pi\colon S \to \Delta(\Acteos)$,
  it is equipped with a mapping
  $f\colon S \to \mathbb{N}^\Acteos$.
Intuitively, for state $s \in S$ and activity $\sigma \in \Acteos$,
  frequency $f(s)(\sigma)$ is the number of times
  $\sigma$ was observed while being in state $s$ (i.e., after parsing a sequence
  ending up in $s$).
In particular, $f(s)(\eos)$ is the number of times a sequence stopped in $s$.
An example FDFA is depicted in Figure~\ref{fig:3gram-fdfa}. For the initial (topmost) state $s$, we have $f(s)(\eos) = 0$,
$f(s)(a) = 13$, and
$f(s)(b) = 17$.

For each such FDFA, we obtain a corresponding PDFA with identical states and transition function and
  $\pi$ defined by $\pi(s)(\sigma) = f(s)(\sigma)  / \sum_{\tau \in \Acteos} f(s)(\tau)$.
Here, we assume $\sum_{\tau \in \Acteos} f(s)(\tau) \ge 1$ for all $s \in S$ which is without restriction
  for the automata constructed via learning.
Function $\pi$ is usually constructed on demand for a given state when a prediction is due or an activity is
  to be sampled.
To simplify notation, we will use an FDFA and its corresponding PDFA synonymously.

We can now express the classical language models within this framework.
A \emph{frequency prefix tree} (FPT) is an FDFA that organizes a finite set of sequences into a tree structure,
  where prefixes of arbitrary length are identified with states (\figref{fig:fpt}).
For $n \ge 1$, an $n$-gram is an FDFA whose states $s$ are identified with the last $n-1$ activities called the \emph{access string} of $s$, i.e., the suffix of length $n-1$ for words of length at least $n-1$ and the word itself for shorter words (see \figref{fig:3gram-fdfa} for a 3-gram).
For example, starting from an initial state (the root in \figref{fig:3gram-fdfa}), the two sequences $ab$ and $aaab$ share the same suffix $ab$ of length $n-1 = 2$, and lead to the same state (the one with $\eos$-frequency 1).
The access string of this state is $ab$.
A \emph{bag} keeps track of the set of activities seen until now.
It is thus an FDFA whose states are subsets of the set~$\Sigma$ of possible activities.

\newcommand{\scaleboxfactor}{0.7}
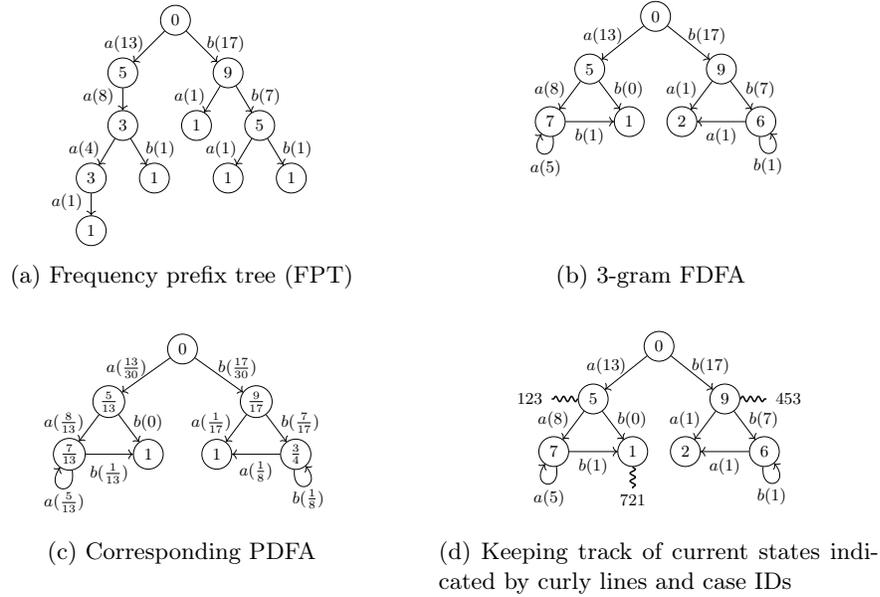
\begin{figure}[thb]
    \centering

    \begin{subfigure}[t]{0.48\textwidth}
        \centering
        \scalebox{\scaleboxfactor}{
        \begin{tikzpicture}[level distance=1.0cm,
        level 1/.style={sibling distance=2cm},
        level 2/.style={sibling distance=1.2cm},
        edge from parent/.style={draw, ->, pos=0.40, inner sep=2.5pt}] 
        \tikzstyle{every node}=[circle,draw]
        
        \node (Root) {0}
            child {
            node (A) {5} 
            child { node (L1) {3}
            		child { node {3}
        					child { node {1} edge from parent node[left,draw=none] {$a(1)$} } 
        					edge from parent node[left,draw=none] {$a(4)$} }    
        			child { node {1} edge from parent node[right,draw=none] {$b(1)$} }
            		edge from parent node[left,draw=none] {$a(8)$} }
            edge from parent node[left,draw=none] {$a(13)$}
        }
        child {
            node (B) {9} 
            child { node (L3) {1} edge from parent node[left,draw=none] {$a(1)$} }
            child { node (L4) {5}
        			child { node {1} edge from parent node[left,draw=none] {$a(1)$} }    
        			child { node {1} edge from parent node[right,draw=none] {$b(1)$} }
            		edge from parent node[right,draw=none] {$b(7)$} }
            edge from parent node[right,draw=none] {$b(17)$}
        };
        
        \end{tikzpicture}
        }
        \caption{Frequency prefix tree (FPT)}
        \label{fig:fpt}
    \end{subfigure}
    \hfill
    \begin{subfigure}[t]{0.48\textwidth}
        \centering
        \scalebox{\scaleboxfactor}{
        \raisebox{1cm}{
        \begin{tikzpicture}[level distance=1.0cm,
        level 1/.style={sibling distance=2.5cm},
        level 2/.style={sibling distance=1.5cm},
        edge from parent/.style={draw, ->, pos=0.30, inner sep=2.5pt}] 
        \tikzstyle{every node}=[circle,draw]
        
        \node (Root) {0}
            child {
            node (A) {5} 
            child { node (L1) {7} edge from parent node[left,draw=none] {$a(8)$} }
            child { node (L2) {1} edge from parent node[right,draw=none] {$b(0)$} }
            edge from parent node[left,draw=none] {$a(13)$}
        }
        child {
            node (B) {9} 
            child { node (L3) {2} edge from parent node[left,draw=none] {$a(1)$} }
            child { node (L4) {6} edge from parent node[right,draw=none] {$b(7)$} }
            edge from parent node[right,draw=none] {$b(17)$}
        };
        
        \draw [->] (L1) to [out=280,in=235,looseness=6] node[below, xshift=0.5mm, yshift=2.2mm, draw=none] {$a(5)$} (L1);
        \draw [->] (L4) to [out=265,in=310,looseness=6] node[below, xshift=-0.5mm, yshift=2.2mm, draw=none] {$b(1)$} (L4);
        
        \draw[->] (L1) to node[below, yshift=2mm, draw=none] {$b(1)$} (L2);
        \draw[->] (L4) to node[below, yshift=2mm, draw=none] {$a(1)$} (L3);
        
        \end{tikzpicture}}}
        \caption{3-gram FDFA}
        \label{fig:3gram-fdfa}
    \end{subfigure}
    ~\\\vspace{0.5cm}
    \begin{subfigure}[t]{0.48\textwidth}
        \centering
        \scalebox{\scaleboxfactor}{
        \begin{tikzpicture}[level distance=1.0cm,
        level 1/.style={sibling distance=2.8cm},
        level 2/.style={sibling distance=1.5cm},
        edge from parent/.style={draw, ->, pos=0.30, inner sep=2.5pt}] 
        \tikzstyle{every node}=[circle, draw]
        
        \node (Root) {0}
            child {
            node[draw, inner sep=1pt] (A) {$\tfrac{5}{13}$} 
            child { node[draw, inner sep=1pt] (L1) {$\tfrac{7}{13}$} edge from parent node[left,draw=none] {$a(\tfrac{8}{13})$} }
            child { node (L2) {1} edge from parent node[right,draw=none] {$b(0)$} }
            edge from parent node[left,draw=none] {$a(\tfrac{13}{30})$}
        }
        child {
            node[draw, inner sep=1pt] (B) {$\tfrac{9}{17}$} 
            child { node (L3) {1} edge from parent node[left,draw=none] {$a(\tfrac{1}{17})$} }
            child { node[draw, inner sep=1.5pt] (L4) {$\tfrac{3}{4}$} edge from parent node[right,draw=none] {$b(\tfrac{7}{17})$} }
            edge from parent node[right,draw=none] {$b(\tfrac{17}{30})$}
        };
        
        \draw [->] (L1) to [out=280,in=235,looseness=6] node[below, xshift=0.5mm, yshift=1.5mm, draw=none, inner sep=0pt, text height=1.5ex, text depth=0.25ex] {$a(\tfrac{5}{13})$} (L1);
        \draw [->] (L4) to [out=265,in=310,looseness=6] node[below, xshift=0.5mm, yshift=1mm, draw=none, inner sep=0pt, text height=1.5ex, text depth=0.25ex] {$b(\tfrac{1}{8})$} (L4);
        
        \draw[->] (L1) to node[below, yshift=2mm, draw=none] {$b(\tfrac{1}{13})$} (L2);
        \draw[->] (L4) to node[below, yshift=2mm, draw=none] {$a(\tfrac{1}{8})$} (L3);
        \end{tikzpicture}
        }
        \caption{Corresponding PDFA}

        \label{fig:3gram-pdfa}
    \end{subfigure}
    \hfill
    \begin{subfigure}[t]{0.48\textwidth}
        \centering
        \scalebox{\scaleboxfactor}{
        \begin{tikzpicture}[level distance=1.0cm,
        level 1/.style={sibling distance=2.5cm},
        level 2/.style={sibling distance=1.5cm},
        edge from parent/.style={draw, ->, pos=0.30, inner sep=2.5pt}] 
        \tikzstyle{every node}=[circle,draw]
        
        \node (Root) {0}
            child {
            node (A) {5} 
            child { node (L1) {7} edge from parent node[left,draw=none] {$a(8)$} }
            child { node (L2) {1} edge from parent node[right,draw=none] {$b(0)$} }
            edge from parent node[left,draw=none] {$a(13)$}
        }
        child {
            node (B) {9} 
            child { node (L3) {2} edge from parent node[left,draw=none] {$a(1)$} }
            child { node (L4) {6} edge from parent node[right,draw=none] {$b(7)$} }
            edge from parent node[right,draw=none] {$b(17)$}
        };
        
        \draw [->] (L1) to [out=280,in=235,looseness=6] node[below, xshift=0.5mm, yshift=2.2mm, draw=none] {$a(5)$} (L1);
        \draw [->] (L4) to [out=265,in=310,looseness=6] node[below, xshift=-0.5mm, yshift=2.2mm, draw=none] {$b(1)$} (L4);
        
        \draw[->] (L1) to node[below, yshift=2mm, draw=none] {$b(1)$} (L2);
        \draw[->] (L4) to node[below, yshift=2mm, draw=none] {$a(1)$} (L3);
        
        \node[left=0.5cm of A, draw=none] (Label) {$123$};
        \draw[decorate, decoration={snake, amplitude=0.5mm, segment length=1.5mm}, thick] (A) -- (Label);
        
        \node[right=0.5cm of B, draw=none] (Label) {$453$};
        \draw[decorate, decoration={snake, amplitude=0.5mm, segment length=1.5mm}, thick] (B) -- (Label);
        
        \draw[decorate, decoration={snake, amplitude=0.5mm, segment length=1.5mm}, thick] 
            (L2) -- ++(0, -0.7) 
            node[midway, below, yshift=0mm, draw=none] {$721$};
        
        \end{tikzpicture}
        }
        \caption{Keeping track of current states indicated by curly lines and case IDs}
        \label{fig:3gram-states}
    \end{subfigure}
    
    \caption{Automata obtained for the event log given by the set
	$L = \{
	    \langle a \rangle^5,
        \langle aa \rangle^3,
        \langle aaa \rangle^3,
        \langle aab \rangle^1,
        \langle aaaa \rangle^1,        
		\langle b \rangle^9,
		\langle ba \rangle^1,
		\langle bb \rangle^5,
		\langle bba \rangle^1,
		\langle bbb \rangle^1
    \}$, with multiplicities of a sequence denoted by powers.
    States (nodes) and transitions upon activities (arrows) are shown.
    Frequencies, respectively, probabilities for activities are indicated in brackets next to activities and within nodes for the $\eos$ activity.
    (a) FPT
	(b) FDFA for the 3-gram, as well as
    (c) PDFA for the 3-gram.
	(d) FDFA for the 3-gram enriched with the current states of cases 123, 453, and 721
    as maintained during inference and streaming learning.
    }
    \label{fig:mainfigure}
\end{figure}

\subsection{Batch Learning}

We next define the prediction problem for batch learning and then adapt the solutions to the streaming case.

In batch learning, we are given an event log
$L \subseteq (\CaseIDs \times \Act)^\ast$,
which is a finite set of finite sequences of case ID-activity pairs.
The goal is to train a language model on $L$, i.e., to construct a prediction function $\Pred \colon \Sigma^\ast \to \Delta(\Acteos)$ that generalizes well and enables informed predictions based on
  event data not necessarily seen before.
Recall that, in our setting, a prediction for a given case ID only
depends on the projection of the event log to this case.
Consequently, it is sufficient to maintain the event order of an event log $L$
  within a case, rather than across all cases.
This allows us to represent $L$ as a finite multiset over $\Sigma^\ast$, i.e.,
  as a mapping $L\colon \Sigma^\ast \to \mathbb{N}$ with finite support.
An example of $L$ over the activity set $\Act = \{a, b\}$ is $L = \{
	    \langle a \rangle^5,
        \langle aa \rangle^3,
        \langle aaa \rangle^3,
        \langle aab \rangle^1,
        \langle aaaa \rangle^1,        
		\langle b \rangle^9,
		\langle ba \rangle^1,
		\langle bb \rangle^5,
		\langle bba \rangle^1,
		\langle bbb \rangle^1
    \}$ with powers denoting the multiplicity within the set.
Here, $L(\varepsilon) = 0$, $L(a) = 5$, $L(bb) = 5$, etc. The multiset contains 30 sequences (cases).

Most batch automata learning algorithms conceptually proceed in two steps \cite{delaHiguera2010}.
First, the event log is arranged as an FPT (\figref{fig:fpt}).
In a second step, this FPT is reduced (or, folded) by identifying, i.e., merging some of the states.
The specificity of an algorithm lies in the second step:
\begin{description}
\item[FPT.] The simplest algorithm directly uses the FPT (and the derived PDFA) as is,
  without any state equivalences.
  While this approach does not generalize well, particularly because no prediction can be made for unseen prefixes, it serves as a baseline and a potential component for ensemble methods.

\item[Alergia.] A commonly used merging algorithm is Alergia \cite{CarrascoO94}.
  The algorithm successively merges nodes in a top-down fashion, with each merge consisting of two stages.
  In the first stage, the algorithm identifies two states that can be merged, typically using a statistical test such as Hoeffding’s incompatibility test \cite{CarrascoO94,HigueraT00}.
  Two states $s$ and $s'$ are considered equivalent (i.e., ``mergeable'') if their probability distributions $p(s)$ and $p(s')$ do not differ significantly, and if their respective successors are equivalent, too.
  In the second stage, the algorithm performs the actual merging of the states and resolves any non-determinism that arises by merging subsequent states and transitions.

\item[Gram.] Finally, an FPT can be reduced to obtain an $n$-gram.
  For this purpose, two states are considered equivalent if the sequence of the last $n-1$ activities in their corresponding access strings is identical (Figures \ref{fig:3gram-fdfa} and \ref{fig:3gram-pdfa}).
  Intuitively, the difference between state merging in Algeria and $n$-grams is that Algeria considers the potential future and $n$-grams the past of a sequence to decide if two states are equivalent.
\end{description}
To optimize computational complexities, implementations, e.g., of $n$-grams, directly construct the FDFA without detouring through an FPT.
We detail this in Section~\ref{sec:streaming} when discussing streaming learning.

Further, we assumed that $L$ is given as a multiset of cases.
While this may be the case for batch learning, inference in production will likely have to cope with
  interleaved cases.
If so, one simply tracks, for every active case ID, the current state in the FDFA (\figref{fig:3gram-states}).

Finally, we note that an LSTM which receives as input a sequence of activities of a case and as output provides a probability distribution over the case's next activities, can be viewed as a probabilistic automaton (with a potentially large number of states).

\subsection{Streaming Learning}
\label{sec:streaming}

In streaming learning, one is given an event stream, where cases are interleaved, rather than an event log, and learning and prediction alternate.
Formally, an event stream is a finite or infinite sequence of events in $\CaseIDs \times \Act$.
While processing the stream, after any activity, we aim to predict the next activity for each
  active case.
For that purpose, after each activity, we perform an update of the automaton and query
  it to return a distribution over the next activities.
For an $n$-gram, the steps are:
\begin{description}
    \item[Initialization.] The FDFA has a single root node $\rho$ corresponding
      to the access string $\varepsilon$ and we set the current state of each (potential) case to $\rho$.

    \item[Update.] Upon receiving an event $(\caseid_i, \act_i)$, with $i \geq 1$,
      from the event stream, the FDFA is updated as follows.
    Let $s$ be the current state of case $\caseid_i$.
    If the case $\caseid_i$ has not been encountered before, i.e., the current state of the     case is $\rho$,
      increment $f(s)(\eos)$.
    We then decrement $f(s)(\eos)$, as one fewer sequence stops
      in $s$, increment $f(s)(\act_i)$, as one more sequence
      continues with $\act_i$, and increment $f(\delta(s, \act_i))(\eos)$,
      provided $\delta(s, \act_i)$ exists.
    We also update the current state of case $\caseid_i$ to $s' = \delta(s, \act_i)$.
    Observe that $f(s')(\eos) > 0$ at the end of the update step,
      so that a potential subsequent decrement does not result in a negative
      $f(s')(\eos)$.
    
    An example is given in \figref{fig:3gram-states}:
    Assume a new event $(123, a)$ is received.
    Case $123$ has been encountered before, and we let $s$ be its current state.
    After the update step, $f(s)(\eos) = 4$, $f(s)(a) = 9$, and
      $f(\delta(s, a))(\eos) = 8$.
    If, on the other hand, $\delta(s, a)$ is not defined,
      a new transition is created (and possibly a new state).
    To illustrate this, consider again \figref{fig:3gram-states}
      and suppose an event $(721, a)$ arrives.
    As the current state of $721$ has the access string $ab$,
      the last two activities seen were $ab$, and so the new
      history of length 2 becomes $ba$.
    A new transition is created to the state with access string $ba$
      and the the counts are adjusted as described above.

    \item[Query.] Upon a query for $\caseid \in \CaseIDs$, the probability distribution induced by $f(s)$ at the current state $s$ of case $\caseid$ is returned.
\end{description}
Both steps are local and computationally efficient on automata as a data structure.
By contrast to the updates in $n$-gram, this locality is not ensured by Alergia.

\subsection{Ensemble Methods}

Ensemble methods \cite{Zhou2012} combine multiple base models, such as FPTs or $n$-grams, into more complex models using voting mechanisms. Voting mechanisms are distinguished based on how they handle updates and inference queries. In both \emph{soft voting} and \emph{hard voting}, each base model individually processes updates from incoming events. Similarly, queries are forwarded to all base models, each of which produces a prediction in the form of a probability distribution. The key difference lies in how these distributions are aggregated: In soft voting, the final probability distribution is obtained by averaging all individual distributions, while in hard voting (also referred to as \emph{majority voting}), each probability distribution is converted into a concrete activity, and the activity chosen most frequently constitutes the final verdict (which may be represented as a Dirac probability distribution).

While hard and soft voting apply to both batch and streaming modes, \emph{adaptive voting} is specific to streaming mode.
The idea is to track the current accuracy of each base model.
A query is then forwarded to the model with the highest accuracy. An update on an incoming event involves two steps. In step 1, the accuracy is adjusted: the activity $\act$ of the event is compared to the prediction of each base model. Base models that correctly predict $\act$ see their accuracy increase, while the accuracy of all other models decreases. Step 2 updates the base models in the same manner as in soft and hard voting.

Finally, we use ensemble methods as a \emph{fallback method} to compensate for slowly-learning language models in streaming learning.
For example, we apply the FPT with a minimum number of 10 visits\footnote{This is a hyperparameter that can be adjusted according to the required confidence level needed.} whenever possible. If the minimum number has not been reached, as a fallback option, the prediction of an $n$-gram is returned, where, for our experiments, we use $n=5$ or $n=7$ depending on the dataset.

We will show in Section \ref{sec:experiments} that soft voting generally outperforms hard voting. Most importantly, soft voting offers considerable improvements over automata-based base models and, in streaming mode, even outperforms LSTMs on many datasets.

\section{Computational Framework}
\label{sec:framework}
A wide range of stream-processing frameworks is available, with examples being Dataflow~\cite{Dataflow}, Faust~\cite{Faust}, Flink~\cite{Flink}, Pathway~\cite{Pathway}, as well as River~\cite{River}.
Towards the goal of a combined batch and stream processing with easy general-purpose function modules, local deployment, analysis via dashboards, low overhead and latencies, and stop-resume functionalities, we developed the \texttt{logicsponge} Python library~\cite{logicsponge}.

A data pipeline in \texttt{logicsponge} is made up of terms that are connected via sequential or parallel composition.
We distinguish two basic types of terms:
source terms and function terms.
The output of a source term does not explicitly depend any other terms in the pipeline.
On the other hand, a function term receives data from other terms as input and can produce output to other downstream terms.
A function term that does not produce any outputs for other terms is a sink.
It may, of course, produce outputs not modeled in the pipeline, like appending data to files or user console output.
In the sequential composition \texttt{t1 * t2} of terms~\texttt{t1} and~\texttt{t2}, the input of~\texttt{t2} is equal to the output of~\texttt{t1}.
In their parallel composition \texttt{t1 | t2}, both~\texttt{t1} and~\texttt{t2} receive the same upstream input and its downstream output is the union of outputs of~\texttt{t1} and~\texttt{t2}, accessible as separate streams.
Function terms that merge these streams into single streams are provided by the framework.
Independently of where they appear in the pipeline, all terms are executed in separate threads.

Inputs and outputs of terms are handled via data streams.
A data stream encapsulates a list of data items.
A data item is a Python dictionary with some associated metadata.
Each data stream is owned by a unique term to which it is associated as an output.
The owner of a data stream can append data items to it.
A data stream can be the input to other terms as defined by the sequential and parallel composition.
Terms that have a data stream as an input do not have write access to it and can access it only via a read-only data stream view object.
The underlying list data structure of a data stream is protected from concurrent access via readers-writer locks~\cite{Raynal2012}, minimizing conflicts.
When starting a pipeline, setting an optional \texttt{persistent} function argument configures the contents of the data streams to be stored in a Python object database~\cite{ZODB}.
Restarting the pipeline after an interrupt or crash restores the state from the database and resumes computation.
The library contains predefined terms to generate plots, compute basic statistics, and create interactive web dashboards.

We implemented the process-mining algorithms used for this work as function terms in the \texttt{logicsponge-processmining} package~\cite{logicsponge-processmining}.
We first define a list of language models:
\begin{pythoncode}
models=[
    BasicMiner(algorithm=Bag()),
    BasicMiner(algorithm=FrequencyPrefixTree()),
    BasicMiner(algorithm=NGram(window_length=2)),
    BasicMiner(algorithm=NGram(window_length=3)),
]
\end{pythoncode}
This list is then provided to a term that performs soft voting on the models:
\begin{pythoncode}
soft_voting = StreamingActivityPredictor(
    strategy=SoftVoting(models=models)
)
\end{pythoncode}
Additionally, we create a term that acts as an LSTM model:
\begin{pythoncode}
lstm = StreamingActivityPredictor(
    strategy=NeuralNetworkMiner(
        model=LSTMModel(vocab_size, *dim, device=device),
        criterion=nn.CrossEntropyLoss(),
        optimizer=optim.Adam(model.parameters(), lr=0.001),
        batch_size=8,
    )
)
\end{pythoncode}
Both terms are combined in parallel with a subsequent respective evaluation term that generates
  statistics on their accuracy and performance.
The pipeline's source term generates a data stream out of the dataset.
The data stream is subsequently processed by applying a key filter and adding start symbol before it
  is fed into the parallel predictor terms:
\begin{pythoncode}
pipeline = (
    ListStreamer(data_list=dataset)
    * ls.KeyFilter(keys=["case_id", "activity"])
    * AddStartSymbol()
    * (
        (soft_voting * Evaluation("soft_voting"))
        | (lstm * Evaluation("lstm"))
    )
)
\end{pythoncode}
Finally, the pipeline is started:
\begin{pythoncode}
pipeline.start()
\end{pythoncode}

\section{Experimental Evaluation}
\label{sec:experiments}

To evaluate our framework and approach, we performed experiments based on seven well-known real-world business-process datasets (Table~\ref{table:dataset-stats}).
All experiments were performed on an 11th Gen Intel Core i9-11900K architecture (3.50GHz, 8 cores, 24GB RAM) with an NVIDIA GeForce RTX 3090 (24GB) running Ubuntu (24.04.1) and Python (3.12.3).
For the LSTM training and inference we used PyTorch with CUDA.
We measured accuracy in both batch and streaming modes. In both modes, we used incremental inference based on the current state of a case.  
In our comparison, we included $n$-grams with $n$ ranging from 1 to 8, FPTs, bags, and LSTMs.  
The LSTMs consisted of two hidden layers of dimension 128, preceded by an embedding layer of dimension 50. 
Every voting method used five submodels, including one FPT, one bag, and three $n$-grams with varying combinations of $n$ (Table~\ref{table:experiments}).
The fallback method is instantiated with an FPT
(considering only states with at least 10 total visits) and an $n$-gram with $n=5$ or $n=7$.

\begin{table}[tbp]
\centering
\caption{Dataset statistics and experimental setup (voting uses FPT + bag + the three $n$-grams specified below)}
\label{table:dataset-stats}
\scalebox{0.87}{
\begin{tabular}{@{}l
>{\hspace{5pt}}c<{\hspace{5pt}}
>{\hspace{5pt}}c<{\hspace{5pt}}
>{\hspace{5pt}}c<{\hspace{5pt}}
>{\hspace{5pt}}c<{\hspace{5pt}}
>{\hspace{5pt}}c<{\hspace{5pt}}
>{\hspace{5pt}}c<{\hspace{5pt}}
>{\hspace{5pt}}c
}
\toprule%
 & \shortstack[c]{Sepsis\\[-0.5ex]Cases\\\cite{sepsis_cases}}
 & \shortstack[c]{BPI\\2012\\\cite{bpi_2012}}
 & \shortstack[c]{BPI\\2013\\\cite{bpi_2013}}
 & \shortstack[c]{BPI\\2014\\\cite{bpi_2014}}
 & \shortstack[c]{BPI\\2017\\\cite{bpi_2017}}
 & \shortstack[c]{BPI\\2018\\\cite{bpi_2018}}
 & \shortstack[c]{BPI\\2019\\\cite{bpi_2019}}
 \\\midrule
 \#Activities & 16 & 24 & 13 & 39 & 26 & 41 & 42		\\
 \#Cases & 1,050 & 13,087 & 7,554 & 46,616 & 31,509 & 43,809 & 251,734\\
 Avg.\ case length & 14.49 & 20.04 & 8.68 & 10.01 & 38.16 & 57.39 & 6.34\\
 \#Events & 15,214 & 262,200 & 65,533 & 466,737 & 1,202,267 & 2,514,266 & 1,595,923\\
 \midrule
 Voting:\\
 (FPT+bag+...) &
 3, 4, 5 &
 3, 5, 7 &
 3, 4, 5 &
 3, 4, 5 &
 3, 5, 7 &
 3, 5, 7 &
 3, 5, 7
 \\[1ex]
 Fallback:\\
 FPT $\to$ ...&
 5-gram &7-gram &5-gram & 5-gram & 7-gram & 7-gram & 7-gram
 \\\bottomrule
 \end{tabular}
 }
\end{table}

\newcommand{\meanacc}{Acc.~(\%)}
\newcommand{\nstates}{\#States}
\newcommand{\latency}{\shortstack[c]{Latency \\[-0.5ex] (ms)}}

\begin{sidewaystable}[tbp]
\centering
\caption{Comparison of batch and streaming learning across datasets}

\scalebox{0.8}{
\begin{tabular}{@{}l >{\hspace{5pt}}rr<{\hspace{5pt}} | >{\hspace{5pt}}rr<{\hspace{5pt}} | >{\hspace{5pt}}rr<{\hspace{5pt}} | >{\hspace{5pt}}rr<{\hspace{5pt}} | >{\hspace{5pt}}rr<{\hspace{5pt}} | >{\hspace{5pt}}rr<{\hspace{5pt}} | >{\hspace{5pt}}rr<{\hspace{5pt}}}

\toprule%
 & \multicolumn{2}{c}{{{\bfseries Sepsis Cases}}}
 & \multicolumn{2}{c}{{{\bfseries BPI 2012}}}
 & \multicolumn{2}{c}{{{\bfseries BPI 2013}}}
 & \multicolumn{2}{c}{{{\bfseries BPI 2014}}}
 & \multicolumn{2}{c}{{{\bfseries BPI 2017}}}
 & \multicolumn{2}{c}{{{\bfseries BPI 2018}}}
 & \multicolumn{2}{c}{{{\bfseries BPI 2019}}}
 \\\toprule
 Batch &
 \meanacc &
 \nstates &
 \meanacc &
 \nstates &
 \meanacc &
 \nstates &
 \meanacc &
 \nstates &
 \meanacc &
 \nstates &
 \meanacc &
 \nstates &
 \meanacc &
 \nstates
 \\\midrule
          	FPT &			43.26 &	4935 &	68.63 &	45277 &	54.01 &	13861 &	33.01 &	152402 &	70.32 &	206498 &	37.71 &	1038724 &	65.75 &	187421\\
\myrowcolor	bag  &        	56.61 &	157 &	73.89 &	319 &	55.79 &	299 &	43.09 &	8994 &		64.25 &	445 &		41.21 &	2598 &		71.85 &	2545\\
		    1-gram  &      20.86 &	1 &		20.15 &	1 &		41.50 &	1 &		17.31 &	1 &			17.18 &	1 &			18.26 &	1 &			17.02 &	1\\
\myrowcolor 2-gram  &      55.90 &	17 &	65.08 &	25 &	60.58 &	14 &	38.44 &	40 &		66.87 &	27 &		57.00 &	42 &		60.10 &	43\\
			3-gram  &      57.78 &	127 &	81.70 &	147 &	70.42 &	92 &	47.52 &	768 &		74.23 &	199 &		63.03 &	621 &		72.33 &	644\\
\myrowcolor	4-gram  &      59.76 &	472 &	83.66 &	420 &	70.71 &	410 &	49.10 &	6003 &		83.71 &	684 &		68.63 &	3664 &		74.00 &	3681\\
			5-gram  &      61.81 &	1250 &	84.82 &	876 &	70.39 &	1267 &	48.50 &	25516 &		87.04 &	1682 &		71.90 &	13673 &		74.46 &	12577\\
\myrowcolor	6-gram  &      59.85 &	2625 &	85.57 &	1552 &	69.79 &	2987 &	46.10 &	69870 &		87.37 &	3422 &		74.86 &	39192 &		74.46 &	29742\\
			7-gram  &      56.20 &	4680 &	85.66 &	2408 &	68.62 &	5778 &	42.43 &	139750 &	87.46 &	6116 &		75.87 &	93049 &		74.36 &	54556\\
\myrowcolor	8-gram  &      52.80 &	7278 &	85.63 &	3576 &	67.16 &	9662 &	38.73 &	225371 &	87.49 &	10003 &		76.34 &	190524 &	74.08 &	83467\\
			fallback     &	61.82 &	N/A &	85.68 &	N/A &	70.55 &	N/A &	48.57 &	N/A &		87.59 &	N/A &		76.59 &	N/A &		74.37 &	N/A\\
\myrowcolor	hard voting  &  64.09 &	N/A	&	85.43 &	N/A &	70.81 &	N/A &	49.75 &	N/A &		86.32 &	N/A &		74.52 &	N/A &		74.76 &	N/A\\
			soft voting  &  64.20 &	N/A &	85.55 &	N/A &	70.99 &	N/A &	49.98 &	N/A &		87.42 &	N/A &		76.36 &	N/A &		74.76 &	N/A\\
\myrowcolor	Alergia  &      50.95 &	47 &	75.25 &	136 &	55.72 &	32 &	31.54 &	74 &		73.48 &	480 &		N/A &	N/A &		70.76 &	373\\
			LSTM  &        	64.00 &	N/A &	85.71 &	N/A &	70.86 &	N/A &	51.90 &	N/A &		88.49 &	N/A &		82.14 &	N/A &		75.27\footnote{only one run and no patience counter } &	N/A
\\\midrule
 Streaming &
 \meanacc &
 \latency &
 \meanacc &
 \latency &
 \meanacc &
 \latency &
 \meanacc &
 \latency &
 \meanacc &
 \latency &
 \meanacc &
 \latency &
 \meanacc &
 \latency
 \\\midrule
FPT &					          	42.84 &	0.04 &	66.39 &	0.06 &	51.09 &	0.04 &	30.97 &	0.09 &	68.92 &	0.04 &	35.90 &	0.04 &	63.90 &	0.14\\
\myrowcolor bag &          			57.95 &	0.09 &	75.86 &	0.06 &	52.68 &	0.04 &	41.67 &	0.07 &	58.43 &	0.05 &	41.69 &	0.04 &	71.58 &	0.05\\
1-gram &         			 		22.21 &	0.13 &	20.91 &	0.06 &	46.14 &	0.06 &	18.96 &	0.30 &	17.42 &	0.05 &	18.44 &	0.05 &	15.70 &	0.07\\
\myrowcolor 2-gram &          		56.57 &	0.09 &	67.78 &	0.06 &	56.34 &	0.04 &	39.97 &	0.06 &	68.66 &	0.06 &	56.80 &	0.05 &	57.92 &	0.05\\
3-gram & 			         		57.76 &	0.08 &	81.37 &	0.06 &	67.17 &	0.04 &	44.97 &	0.07 &	74.19 &	0.04 &	62.94 &	0.04 &	72.00 &	0.05\\
\myrowcolor 4-gram &          		58.98 &	0.08 &	83.34 &	0.07 &	67.19 &	0.04 &	45.85 &	0.06 &	83.79 &	0.06 &	68.55 &	0.04 &	73.81 &	0.05\\
5-gram &          					60.37 &	0.13 &	84.48 &	0.06 &	66.49 &	0.04 &	44.46 &	0.07 &	87.12 &	0.05 &	71.73 &	0.04 &	74.08 &	0.08\\
\myrowcolor 6-gram &	       	   	57.80 &	0.12 &	85.15 &	0.06 &	65.22 &	0.05 &	41.55 &	0.06 &	87.44 &	0.05 &	74.45 &	0.04 &	73.89 &	0.06\\
7-gram &         				 	53.82 &	0.16 &	85.15 &	0.06 &	63.47 &	0.04 &	38.03 &	0.11 &	87.49 &	0.05 &	75.12 &	0.05 &	73.54 &	0.08\\
\myrowcolor 8-gram &				50.18 &	0.08 &	85.05 &	0.07 &	61.46 &	0.05 &	34.99 &	0.06 &	87.46 &	0.05 &	75.15 &	0.05 &	73.16 &	0.05\\
fallback &					 		60.40 &	0.05 &	85.13 &	0.06 &	66.66 &	0.05 &	44.48 &	0.07 &	87.58 &	0.06 &	76.06 &	0.05 &	73.55 &	0.08\\
\myrowcolor hard voting &    	    63.68 &	0.28 &	84.97 &	0.58 &	67.23 &	0.17 &	46.05 &	0.19 &	84.91 &	0.15 &	74.22 &	0.16 &	74.26 &	0.20\\
soft voting &          				64.80 &	0.17 &	85.20 &	0.11 &	67.46 &	0.07 &	46.84 &	0.13 &	87.50 &	0.09 &	74.39 &	0.07 &	74.56 &	0.12\\
\myrowcolor adaptive voting &		60.37 &	0.28 &	85.15 &	0.23 &	67.19 &	0.15 &	45.84 &	0.22 &	87.49 &	0.17 &	75.12 &	0.15 &	74.08 &	0.23\\
LSTM &          					61.66 &	6.41 &	84.34 &	4.46 &	66.33 &	7.00 &	44.39 &	7.83 &	87.27 &	7.21 &	80.15\footnote{stopped after 8 hours and 2,351,500 events} &	10.03 &	75.75\footnote{stopped after 14 hours and 1,471,201 events} &	30.14
\\\bottomrule
\end{tabular}
}

\label{table:experiments}
\end{sidewaystable}

\subsection{Batch Mode}

In batch mode, a given dataset (event log) was divided into
a training set (70\%), a validation set (15\%), and a test set (15\%),
where the percentages refer to the number of sequences in the event log.
To account for $\eos$-predictions, we added a $\eos$ symbol to the end
of each sequence.
The validation set was used only for LSTM training to determine a stopping criterion.
We averaged the results over 5 runs. We performed at most 20 epochs per run with a batch size of 8, but stopped a run as soon as a model failed to improve for 3 consecutive epochs. Per run, we then selected the LSTM model with the best validation accuracy.

Every model was evaluated on the test set.
For every activity $a$ in the dataset, a given model selected
the outcome $\hat{a}$ with the highest probability
based on its current state as its prediction.
Thus, the prediction was correct if $a = \hat{a}$.
Accuracy measured the proportion of correct predictions.

For $n$-grams, we implemented backoff \cite{Katz87}:
if the next activity cannot be parsed by an $n$-gram, the model
attempts to parse shorter and shorter suffixes, always starting
from the initial state. The test results for Alergia were obtained for significance parameter 0.5,
using the automata-learning library AALpy \cite{MuskardinAPPT22}.

\subsection{Streaming Mode}

In streaming mode, we did not add $\eos$-symbols to mark the end
of a sequence, as adding them to pre-mortem data is unsuitable
unless explicitly present in the dataset.
On the other hand, to handle the first event of a case in a data stream, we insert
an ``init'' symbol to mark the start of a sequence
allowing an LSTM to make a prediction from the very beginning.

For querying an LSTM, we parse the current sequence through
the model and retrieve the corresponding outcome.
LSTMs are not inherently designed for streaming learning.
We update a model efficiently with each incoming event as follows:
After each new event, we perform a training iteration.
Each training step uses a batch size of 8, always including
the updated case sequence. The remaining sequences are selected
in a round-robin fashion to ensure all sequences are considered
repeatedly.

Note that, unlike in batch learning, there is no separation of the
dataset in training, validation, and test set.

\subsection{Results}

Experimental results are depicted in Table~\ref{table:experiments}.
In batch mode, LSTMs perform best across most datasets. However, in most cases, an $n$-gram model comes very close. Combining these models, along with frequency prefix trees and bags, into ensemble methods further boosts the performance of simple base models: in two cases, LSTMs are even outperformed (by soft voting).

\begin{figure}[t]
    \centering
    \begin{subfigure}[b]{0.99\textwidth}
        \centering
        \includegraphics[width=\textwidth, trim=0.5cm 1.5cm 1.5cm 1.6cm, clip]{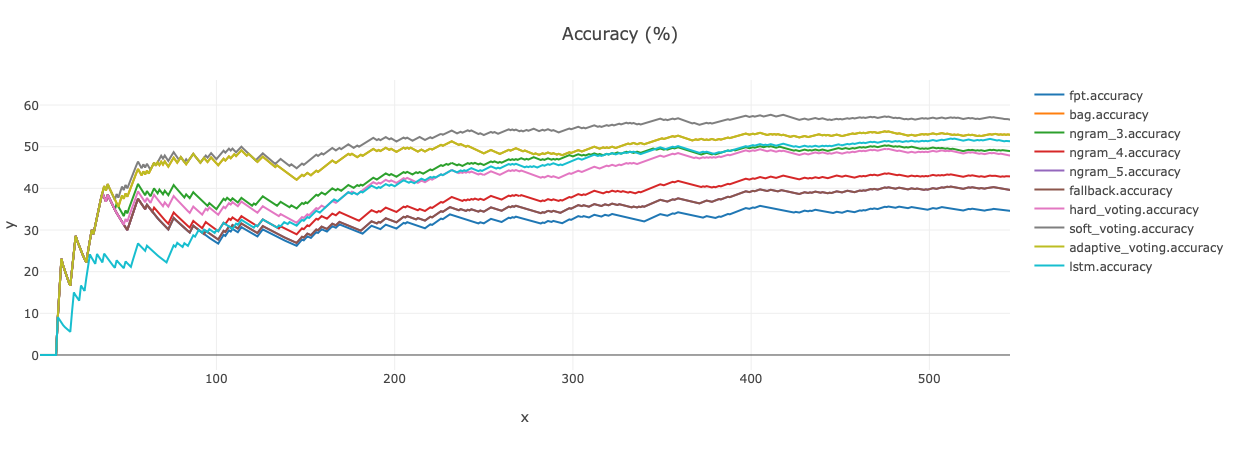}
        \caption{First 550 predictions}
        \label{fig:streaming-initial}
    \end{subfigure}\\
    \begin{subfigure}[b]{0.99\textwidth}
        \centering
        \includegraphics[width=\textwidth, trim=0.5cm 1.5cm 1.5cm 1.6cm, clip]{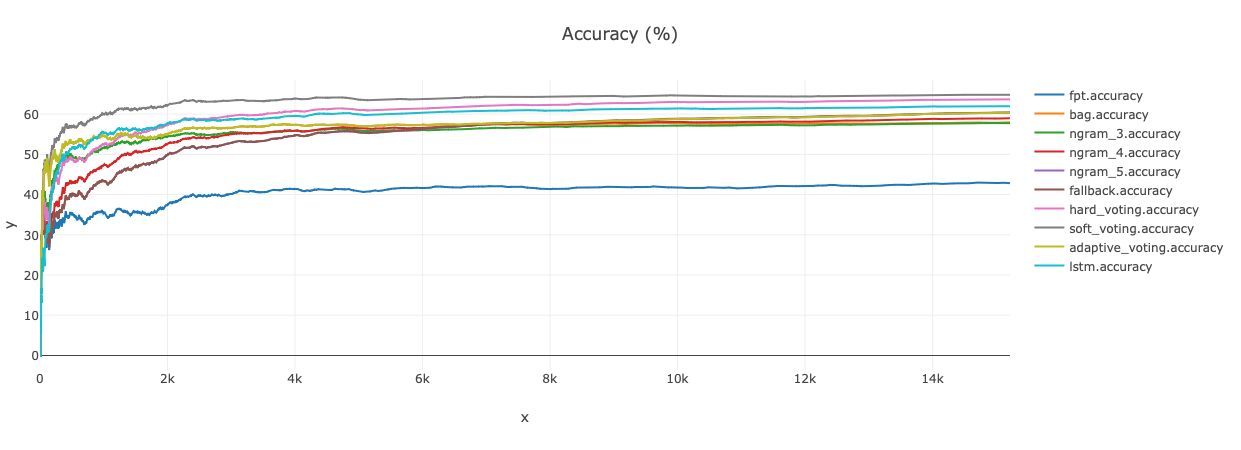}
        \caption{Predictions for the complete dataset}
        \label{fig:streaming-full}
    \end{subfigure}
    \caption{Prediction accuracy for different language models in streaming mode over the number of activities in the sepsis dataset from \cite{sepsis_cases}.}
    \label{fig:streaming}
\end{figure}

In streaming mode, LSTMs appear to struggle during the initial stage
of a stream. As illustrated in \figref{fig:streaming-initial} for the sepsis dataset,
which depicts the accuracy evolution for various models, LSTM performance
is initially well below that of other base models, catching up only later (\figref{fig:streaming-full}). Ensemble methods can  outperform LSTMs
on several datasets. Notably, the average latency
(the total time required for making one prediction and one update) of base models is significantly lower that of LSTMs.

\section{Conclusion and Future Work}
\label{sec:conclusion}

We presented a computational framework for event-log predictions in streaming mode.
Experimental results on business processes carried out in this framework suggest that process data
  often exhibits behavior that is consistent with
  computationally efficient $n$-gram predictions.
Moreover, ensemble methods considerably boost smaller base models, at the expense of only little
  computational overhead.

One direction for future research is refining the methodology.
Fallback algorithms show promising potential, but more detailed experiments are required
  to explore more sophisticated fallback strategies that rely,
  for example, on statistical tests.
A potential starting point is the work in automata learning \cite{delaHiguera2010}) to compare
  states as well as the work on probabilistic bisimilarity distances \cite{Breugel17}.
It would be interesting to investigate whether ensemble methods involving nested models (such as voting among other voting models) could further improve performance. In streaming mode, more fine-grained adaptive methods, particularly concerning LSTM training, may help reduce latency and avoid overfitting.

From a theoretical perspective, it would be worthwhile to explain the $n$-gram phenomenon:
Can a random automaton or process often be approximated by an $n$-gram?
What is the optimal $n$, or what are the best combinations of $n$'s in ensemble methods? Can these be determined online? Is it worthwhile to distribute them sparsely, or are clusters of $n$ preferable?
In other words, what is the best way to determine the optimal hyperparameters?

As object-centric process mining attracts increasing attention \cite{abs-2312-09741},
  incorporating global behaviors into predictions gains in importance.
In terms of automata, a natural model that distinguishes between global
  and local behaviors is data automata \cite{BojanczykDMSS11}, which have already been
  studied in a classification learning setting \cite{DeckerHLT14}.
Frequency-enriched and probabilistic versions are hypothesized to be good starting points to
  pursue this direction beyond the local predictors considered in this paper.
Another extension involves incorporating more attributes or activities, including continuous ones such as energy consumption, clinical parameters, or timestamps and activity durations \cite{AalstSS11,HinkkaLH19}.


\bibliographystyle{abbrv}
\bibliography{lit}

\end{document}